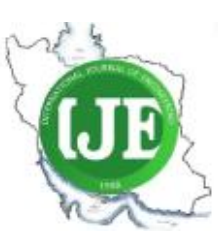

# A New Modeling to Feature Selection Based on the Fuzzy Rough Set Theory in Normal and Optimistic States on Hybrid Information Systems

M. H. Safarpour, S. M. Alavi*, M. Izadikhah, H. Dibachi

*Department of Mathematics and Computer Science, Arak Branch, Islamic Azad University, Arak, Iran*

*ABSTRACT*

Considering the high volume, wide variety, and rapid speed of data generation, investigating feature selection methods for big data presents various applications and advantages. By removing irrelevant and redundant features, feature selection reduces data dimensions, thereby facilitating optimal decision-making within decision systems. One of the key tools for feature selection in hybrid information systems is fuzzy rough set theory. However, this theory faces two significant challenges: First, obtaining fuzzy equivalence relations through intersection operations in high-dimensional spaces can be both time-consuming and memory-intensive. Additionally, this method may produce noisy data, complicating the feature selection process.The purpose and innovation of this paper are to address these issues. We proposed a new feature selection model that calculates the combined distance between objects and subsequently used this information to derive the fuzzy equivalence relation. Rather than directly solving the feature selection problem, this approach reformulates it into an optimization problem that can be tackled using appropriate meta-heuristic algorithms. We have named this new approach **FSbuHD**. The FSbuHD model operates in two modes—normal and optimistic—based on the selection of one of the two introduced fuzzy equivalence relations. The model is then tested on standard datasets from the UCI repository and compared with other algorithms. The results of this research demonstrate that FSbuHD is one of the most efficient and effective methods for feature selection when compared to previous methods and algorithms.

### Graphical Abstract

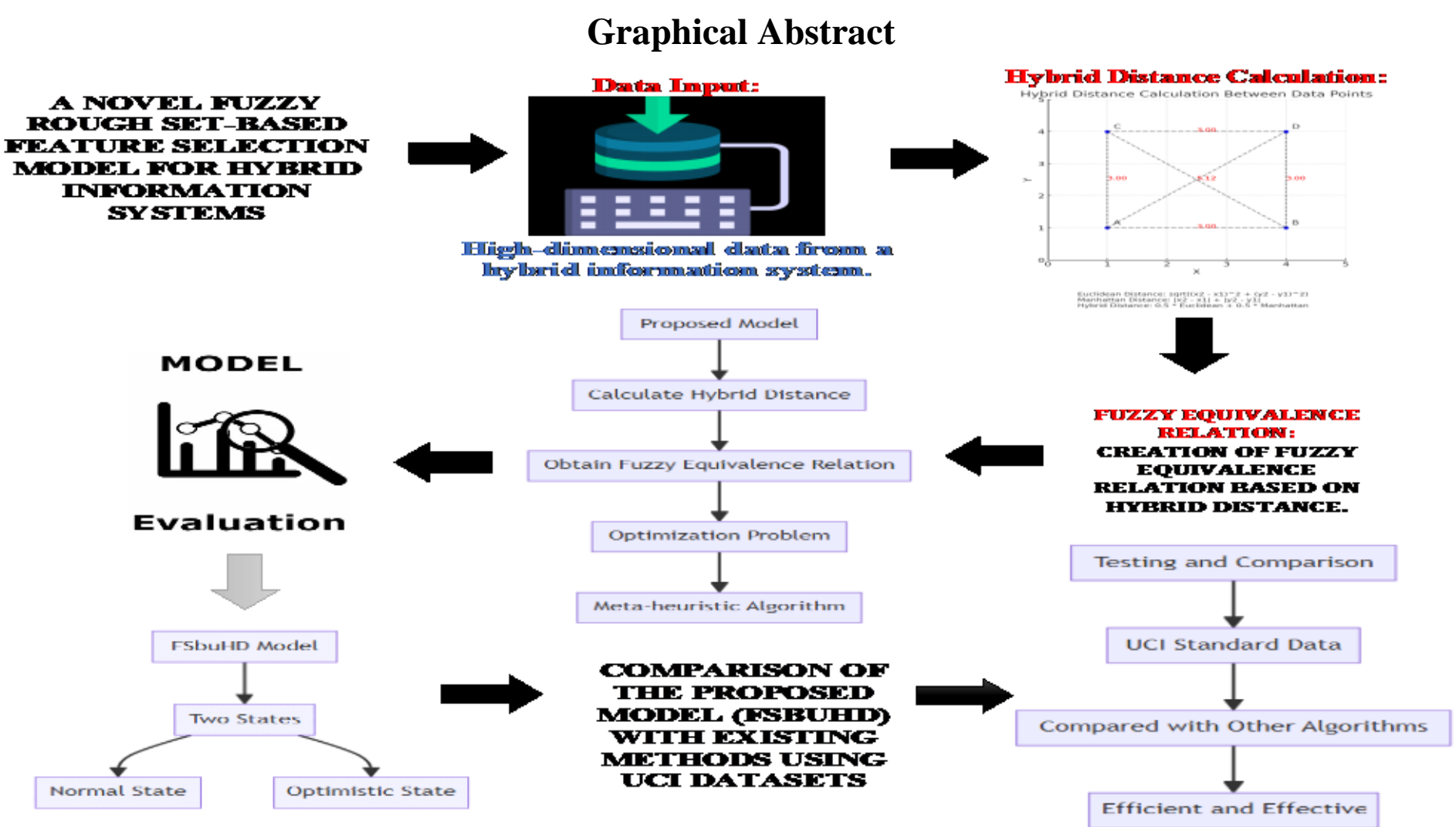

*Corresponding Author Email: sm.alavi@iau.ac.ir (S. M. Alavi)

## 1. INTRODUCTION

In today's world, the exponential growth of data has led to the emergence of big data, characterized by four key attributes: volume, variety, velocity, and value (1-3). The immense volume of data generated encompasses diverse formats, including structured, unstructured, and semi-structured data from numerous sources. Furthermore, the high velocity at which data is produced presents significant challenges for efficient processing and analysis.

Another challenge related to big data is ensuring its quality and reliability, as it is often obtained from external systems, which presents the risk of the data being inaccurate or outdated (4). Traditional production and operation modes from the industrial economy era have been fundamentally impacted by the rapid development of information technology and the Internet. The structure and operation of the network economy have evolved through the integration of data and information, involving a variety of entities, such as equipment, individuals, and enterprises. Big data technology and management tools enable enterprises to integrate information resources, transform data into knowledge, and accelerate knowledge creation (5).

The potential value of big data lies in extracting valuable insights and information through accurate analysis and effective utilization (6). This potential has created a critical need for feature selection algorithms to enhance knowledge discovery across various domains. Feature selection, also known as variable selection or feature subset selection, is the process of selecting subsets of features used in model construction. Datasets often contain numerous redundant or correlated features that can be eliminated without significant information loss before applying feature selection techniques. Feature extraction, in contrast, generates new features from the original ones, producing a transformation of the original features. While the compression of new features can be more efficient, it may compromise the physical interpretability of the original features, which is a limitation (7).

Feature selection is a fundamental technique in pattern recognition, machine learning, and data mining, focusing on identifying and selecting the most relevant and informative features within a dataset (8). Many feature selection methods exist, including filtering, wrapping, and embedding approaches, each with its own advantages, limitations, and applicability to various data types (9).

Fuzzy rough set theory is a valuable tool for feature selection and dimensionality reduction in big data, and it has attracted significant interest. Considerable research has explored different methodologies and algorithms based on this theory. Rough set theory, specifically, provides a unique mathematical framework for identifying and selecting relevant features in datasets, accommodating both qualitative and quantitative attributes. Further examination of rough set theory reveals that many studies use it in conjunction with other computational techniques, such as neural networks or metaheuristic algorithms, thus enriching the research landscape. This interdisciplinary approach not only reflects the robustness of the theory but also highlights a trend toward integrating rough set principles with emerging methodologies in data science to enhance feature selection processes (10). Consequently, rough set theory (RST) has proven to be an effective tool for feature selection, with the capacity to handle multi-source data (11, 12).

However, RST has two major drawbacks: using the similarity relation derived from rough set theory in high-dimensional spaces often leads to noise, requiring considerable time and cost. This paper aims to address these drawbacks by introducing a new fuzzy similarity relation derived from the hybrid distance between objects.

This research focuses on developing a feature selection approach utilizing fuzzy rough sets in hybrid information systems (HIS). The objective is to improve the effectiveness of feature selection by incorporating the distance between elements in HIS, rather than relying solely on similarity relations. Data mining and machine learning (ML) have garnered academic attention worldwide as heuristic strategies yield promising results (13). By employing this approach, the most relevant and informative features within HIS can be identified and selected with greater accuracy and efficiency. This method has the potential to enhance the performance of various learning and recognition tasks by reducing the dataset's dimensionality while preserving its essential characteristics.

The structure of the paper is as follows: Section 2 describes related works. Section 3 introduces the fundamental concepts of fuzzy rough set theory and the computation of hybrid distance. Section 4 presents the feature selection method. Section 5 details a series of experiments performed on publicly available data, with results analyzed and compared. Finally, section 6 summarizes the findings.

## 2. RESEARCH BACKGROUND

The classical rough set, introduced by Pawlak (14), is especially effective for processing symbolic or discrete data, allowing identification and selection of significant and relevant features for further analysis. However, in many practical applications, databases often become large and complex, resulting in a hybrid information system (HIS) that combines multiple attribute types (15, 16). In learning or recognition tasks, certain attributes may be irrelevant, and including these in learning algorithms can degrade performance by increasing dimensionality, leading to longer training and testing times and ultimately reducing system efficiency (17, 18). Consequently, identifying and eliminating irrelevant attributes from the HIS is crucial to improving the accuracy and efficiency of these tasks.

In HIS, real-valued data usually require discretization before feature selection. However, this process risks losing the mutual information contained in the original real-valued features. To address this challenge, Dubois and Prade (19) introduced the classical fuzzy rough set (FRS), which can handle real-valued features directly. In FRS, crisp equivalence relations are replaced with membership degrees that indicate an object's affiliation with decision classes. Over time, several extensions of FRS models have been developed (20, 21), introducing various operators to define fuzzy similarity relations. Examples include the T-similarity measure, the R-implication measure (22), and kernel-based similarity measures (23). The application of FRS models to feature selection has gained considerable attention, leading to numerous studies focusing on implementing FRS-based feature selection techniques (24, 25).

Various attribute reduction algorithms have been developed to identify minimal attribute sets using discernibility matrices, such as the algorithm proposed by Chen et al. (26). In a similar approach, Chen and Yang [24] combined classical rough sets with FRS to define a discernibility relation for each symbolic and real-valued attribute, employing intersection operators to determine attribute relevance.

Wang et al. (27) introduced the parameterized fuzzy relation, which enables the analysis of real-valued data by constructing a fuzzy neighborhood rough set model based on the relationship between fuzzy neighborhoods and fuzzy decisions. Fuzzy similarity relations are the core of fuzzy rough set models, with each model being defined by the specific type of fuzzy similarity relations it uses. Readers may refer to standard reference texts in discrete mathematics for further insights into various methods for establishing fuzzy similarity relations (28). The primary concept underlying these algorithms is the use of intersection operations to calculate fuzzy similarities between samples, which then serves to assess the dependence between attributes and decisions (29). For example, with a subset of attributes BBB and a fuzzy similarity relation $R_B$ induced by $B$, the fuzzy similarity relation $R_B$ is determined by taking the intersection of multiple fuzzy similarity relations, specifically

$R_B = \bigcap_{a \in B} R_a$,

where $R_a$ is a fuzzy similarity relation induced by attribute aaa. However, a potential issue with this approach of calculating the similarity between samples is that it may result in limited discrimination of membership degrees in a fuzzy relation, especially when noise is present in the dataset (30). The effect of noise can be amplified by repeated intersection operations, leading to a less accurate estimation of the fuzzy similarity relation. Revising feature reduction in fuzzy rough sets is thus motivated by the fact that the membership function of a fuzzy similarity relation, generated by intersection operations, may not accurately reflect the relationship between instances (31, 32).

In traditional fuzzy similarity relations, intersection operations are used to compute the degree of similarity between samples. However, this approach has limitations in capturing the true relationship between samples accurately (33). To address this limitation, the use of distance measures is proposed. Distance measures are widely employed to characterize similarity between samples and do not depend on fuzzy sets to compute similarity relations. Therefore, fuzzy similarity relations defined through distance measures can overcome the shortcomings of those generated by intersection operations.

A model for attribute reduction is constructed by introducing a distance measure into fuzzy rough sets to address the weaknesses arising from intersection operations (34). By utilizing distance measures, a more accurate and efficient approach to attribute reduction in fuzzy rough sets can be developed, improving performance in various learning and recognition tasks, such as pattern recognition, machine learning, and data mining. This approach, which incorporates hybrid distance and a new definition of similarity relationships, has led to the construction of a novel model, which is described in detail below.

## 3. PRELIMINARIES

In this section, fundamental concepts related to fuzzy rough sets and the various attributes in hybrid information systems are reviewed. Additionally, calculating distances between these attributes is a critical step in data analysis and will be discussed in detail.

### 3. 1. Fuzzy Rough Set

**Definitin 1**. An information system is defined as a $\langle U, A \cup D, V \rangle$, where $U$ is a non-empty finite set of objects, $A$ is a non-empty finite set of attributes, and $D$ denotes the set of decision attributes, $A \cap D = \emptyset$. Each attribute $a \in A$ or $d \in D$ is associated with a set $V_a$ or $V_d$ of its value, called the domain of $a$ or $d$, respectively (34).

The rough set theory describes a crisp subset of a universe by two definable subsets called lower and upper approximations.

**Definition 2**. Let $U$ be a non-empty finite universe, then $R \subseteq U \times U$ is called an equivalence relation on $U$, if for $\forall x, y, z \in U$, (35)

$$\begin{aligned}&(1)\ Reflexivity: R(x,x)=1,\\&(2)\ Symmetry: R(x,y)=R(y,x),\\&(3)\ Transitivity: R(x,y)=R(y,z)\Rightarrow\ R(x,y)=R(x,z)\end{aligned}\qquad(1)$$

The equivalence relation partitions the universe into a family of disjoint subsets called equivalence classes. $[x]_R$ denotes an equivalence class of $R$ containing an element $x\in U$.

**Definition 3**. Let $U$ be a non-empty finite universe, $R\subseteq U\times U$ is an equivalence relation on $U$, $U/R$ denotes the family of all equivalence classes $R$. For any $X\subseteq U$, the lower and the upper approximations of $X$ are defined as follows, respectively (19):

$$\begin{cases}\underline{R}X=\{x\in U|[x]_R\subseteq X\}\\ \overline{R}X=\{x\in U|[x]_R\cap X\neq\emptyset\}.\end{cases}\qquad(2)$$

**Definition 4**. Let $U$ be a non-empty finite universe, and $A(.): U\to[0,1]$ be a mapping. $A$ is called a fuzzy set on $U$, for any $x\in U$, and $A(x)$ is called membership degree of $x$ to $A$ (36).

**Definition 5**. A fuzzy number $A$ is a fuzzy set of the real line with a normal, convex, and continuous membership function of bounded support (37).

**Definition 6**. A fuzzy set $A$ is called trapezoidal fuzzy number with tolerance interval $[b,c]$, left width $(b-a)$ and right width $(d-c)$ if its membership function has the following form (37, 38):

$$A(t)=\begin{cases}1-(b-t)/b-a, & if\ \ a\leq t\leq c,\\ 1, & if\ \ b\leq t\leq c,\\ 1-(t-c)/d-c, & if\ \ b\leq t\leq d,\\ 0, & if\ \ otherwise.\end{cases}\qquad(3)$$

The notation A = (a, b, c, d) shows a trapezoidal fuzzy number, if b=c then the number fuzzy is called triangular fuzzy number.

**Definition 7**. Let $U$ be a non-empty finite universe, and $R\in F(U\times U)$, $R$ is called a fuzzy equivalence relation on $U$, if for $\forall x,y,z\in U$, (35):

$$\begin{aligned}&(1)\ Reflexivity: R(x,x)=1,\\&(2)\ Symmetry: R(x,y)=R(y,x),\\&(3)\ Min-transitivity: min\{R(x,y),R(y,z)\}\leq R(x,z)\end{aligned}\qquad(4)$$

**Definition 8**. A binary operator $T$ on the unit interval $[0,1]\times[0,1]$ is said to be a triangular norm. If $\forall a,b,c\in[0,1]$, the following conditions are satisfied (39):

$$\begin{aligned}&(1)\ Commutativity: T(a,b)=T(b,a),\\&(2)\ Associativity: T(T(a,b),c)=T(a,T(b,c)),\\&(3)\ Monotonicity: a\leq c,b\leq d\Rightarrow T(a,b)\leq T(c,d),\\&(4)\ Boundary\ \ condition: T(a,1)=a, T(1,a)=a\end{aligned}\qquad(5)$$

As an example, $T_p(a,b)=ab$ is called probability t-norm. A binary operator $S$ on the unit interval $[0,1]\times[0,1]$ is a triangular-conorm (shortly, t-conorm or s-norm) if it is increasing, associative, and commutative and satisfies the boundary condition of the form $S(a,0)=a,\forall a\in[0,1]$.

**Definition 9**. Let $U$ be a non-empty finite universe, and $R\in F(U\times U)$, $R$ is called a T-fuzzy equivalence relation on $U$, if for $\forall x,y,z\in U$,

$$\begin{aligned}&(1)\ Reflexivity: R(x,x)=1,\\&(2)\ Symmetry: R(x,y)=R(y,x),\\&(3)\ T-transitivity: T(R(x,y),R(y,z))\leq R(x,z).\end{aligned}\qquad(6)$$

**Definition 10**. Let $R$ be a fuzzy equivalence relation on $U$ and $X$ be a fuzzy subset on $U$. The fuzzy lower and upper approximations of $X$ is defined as follows, respectively (40):

$$\begin{cases}\underline{R}X(x)=\inf\limits_{y\in U}\left\{S_p\{1-R(x,y),X(y)\}\right\},\\ \overline{R}X(x)=\sup\limits_{y\in U}\left\{T_p\{R(x,y),X(y)\}\right\}.\end{cases}\qquad(7)$$

**Definition 11**. Let $\langle\ U,A\cup D,V\rangle$ be the information system, where $U=\{x_1,x_2,\dots,x_n\}$ is a finite set of objects and $A=\{a_1,a_2,\dots,a_m\}$ is the condition attribute set. $R_a$ is defined a fuzzy relation for each condition attributes $a\in A$, $D=\{d\}$ is the decision attributes, the sample set $U$ is divided into r crisp equivalence classes by $D$, $U/D=\{D_1,D_2,\dots,D_r\}(1\leq j\leq r)$ (19). $B$ is a subset of $A$, denoted as $B\subseteq A$, and $R_a$ is the fuzzy similarity relation for each $a\in B$. Let

$$R_B=\cap_{a\in B}R_a,\qquad(8)$$

Then, $R_B$ is a fuzzy similarity relation on $U$. Furthermore, the membership function of the decision class $D_j$ is:

$$D_j(x)=\begin{cases}1, & x\in D_j,\\ 0, & Otherwise.\end{cases}\qquad(9)$$

To handle the uncertainty of the decision feature and according to the Relation 7, the classical FRS introduces the definitions of the fuzzy rough approximations as follows:

$$\begin{aligned}&\underline{R_B}(D_j)(x)=\inf\limits_{y\notin D_j}\{1-R_B(x,y)\}, D_j\in U/D,\\&\overline{R}_B(D_j)(x)=\sup\limits_{y\in D_j}\{R_B(x,y)\}, D_j\in U/D\end{aligned}\qquad(10)$$

where $\underline{R_B}(D_j)(x)$ and $\overline{R}_B(D_j)(x)$ are called the fuzzy lower and upper approximations of $D_j$, respectively. $\underline{R_B}(D_j)(x)$ denotes that the membership of $x$ certainly belonging to $D_j$ equals to the minimum of the dissimilarities between $x$ and all objects from the sample domain $U-D_j$. $\overline{R}_B(D_j)(x)$ means that the membership of $x$ can belong to $D_j$ equals the maximum of the similarities between all objects of $D_j$.

**Definition 12**. The definition of the fuzzy positive region $POS_B(D)$ of the condition attribute subset concerning the decision attribute $D$ can be expressed as follows (41):

$$POS_B(D) = \cup_{j=1}^{r} \underline{R}_B(D_j) \tag{11}$$

**Definition 13**. The fuzzy dependency function can be calculated using the definition of the fuzzy positive region, as follows:

The ratio of the positive region to all objects in the feature space is defined as $\gamma_B(D)$,

$$\gamma_B(D) = \frac{|POS_B(D)|}{|U|} \tag{12}$$

where $|.|$ denotes the cardinality of a set. It is important to note that $\gamma_B(D)$ falls within the range of 0 to 1, and it is utilized to assess the importance of a subset of features. The fuzzy dependency, conversely, is a measure of the classification ability of a particular feature subset $B$.

**Definition 14**. A hybrid information system is defined as a $HIS = \langle U, A \cup D, V \rangle$, where $U$ is a non-empty finite set of objects, and $A$ is a non-empty finite set and consists of different types of attributes, let $A = A^r \cup A^s \cup A^c \cup A^b \cup A^l$, $A^r$ is the real-valued attribute set, $A^s$ is the set-valued attribute set, $A^c$ is the categorical attribute set, $A^b$ is the boolean attribute set and $A^l$ is the linguistic variable attribute set and, $A^r \cap A^s = \emptyset$, $A^r \cap A^c = \emptyset$, $A^r \cap A^b = \emptyset$, $A^r \cap A^l = \emptyset$, $A^s \cap A^c = \emptyset$, $A^s \cap A^b = \emptyset$, $A^s \cap A^l = \emptyset$, $A^c \cap A^b = \emptyset$, $A^c \cap A^l = \emptyset$, $A^b \cap A^l = \emptyset$, D denotes the set of decision attributes, $A \cap D = \emptyset$, and $V$ shows the domain set of attributes (39).

## 3. 2. Hybrid Distance

To construct an efficient distance function among objects in a Health Information System (HIS), it is essential to develop a hybrid distance function. This necessity arises because HIS may contain various types of attributes, including Boolean, real-valued, set-valued, and linguistic variable attributes. To address the value differences across these diverse attributes, the following definition for value difference is proposed.

The concept of a linguistic variable provides a means to approximately characterize phenomena that are too complex or ill-defined to be adequately described in conventional quantitative terms. Specifically, treating the illness rate as a linguistic variable with values such as "minor," "moderate," and "severe" enables the application of fuzzy logic.

In this research, the linguistic variable attribute distance (LD) is defined in two steps. In the first step, the conversion of the linguistic variable into a trapezoidal fuzzy number is carried out. For instance, the linguistic variable "T(illness rate) = {Minor, Moderate, Severe}" can be represented as a trapezoidal fuzzy number, as illustrated in Figure 1.

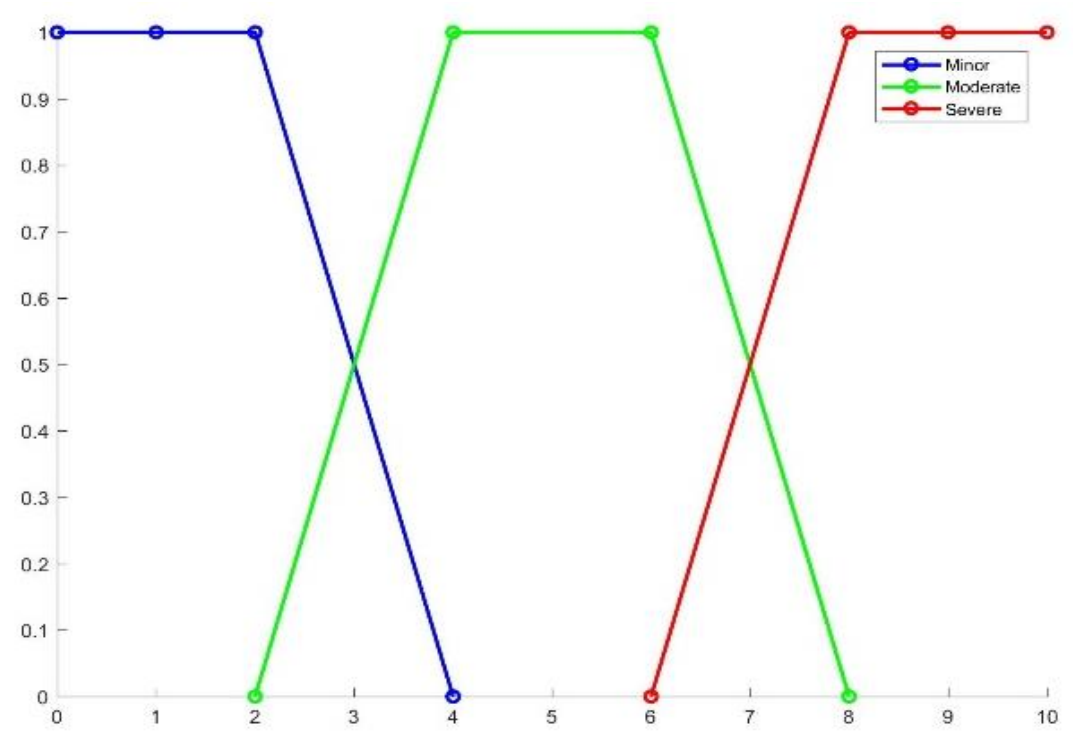

**Figure 1.** Linguistic variable of the illness rate as the trapezoidal fuzzy number

In the second step, defuzzification is performed, which involves converting a collection of fuzzy values or membership functions—known as a fuzzy set—into a single numerical or crisp value. While several methods are available for defuzzification, the centroid method is utilized in this study. This method calculates the crisp value as the center of gravity or the weighted average of the trapezoidal fuzzy number. The centroid method typically strikes a good balance between accuracy and simplicity, making it a widely used approach in various applications.

Assuming that $\mu_{\tilde{N}}$ be a fuzzy set then the formula for the centroid method is expressed as follows (42):

$$c_{\mu_{\tilde{N}}} = \frac{\int x\mu_{\tilde{N}}(x)dx}{\int \mu_{\tilde{N}}(x)dx} \tag{13}$$

In the special case, if A= (a, b, c, d) is a trapezoidal fuzzy number, then the centroid method can be written as follows:

$$c_A = (3a + b + 3c + 2d)/9 \tag{14}$$

**Definition 15**. Let $HIS = \langle U, A \cup D, V \rangle$, for each $x, y \in U$, then $a(x)$, $a(y)$ denote the values of $x$, $y$ under $a$, respectively. Relations 12-15 show the attributes distance in four different cases (43).

1) Let $a$ be a boolean attribute, for each $a \in A$. The boolean attribute distance $(BD)$ is defined as follows:

$$BD(a(x), a(y)) = \begin{cases} 0, & a(x) = a(y), \\ 1, & a(x) \neq a(y). \end{cases} \tag{15}$$

2) Let $a$ be a real-valued attribute, for each $a \in A$. The real-valued attribute distance $(RD)$ is defined as follows:

$$RD(a(x), a(y)) = \frac{|a(x) - a(y)|}{4\sigma_a} \tag{16}$$

where $\sigma_a$ is the standard deviation under the attribute $a$.

3) Let $a$ be a set-valued attribute, for each $a \in A$, and $V_a$ is the domain of $a$. The set-valued attribute distance $(SD)$ is defined as follows:

$$SD(a(x), a(y)) = 1 - \frac{a(x) \cap a(y)}{s} \tag{17}$$

where $s$ is the maximum cardinal number (cardinality) in the subset of $V_a$.

4) Let $a$ be a linguistic variable attribute, for each $a \in A$. The linguistic variable attribute distance $(LD)$ is defined as follows, there are different formulas to find the distance between two fuzzy numbers, and in this study, the following formula is used.

$$LD(a(x), a(y)) = \frac{|c_1 - c_2|}{4\sigma_a} \quad (18)$$

$c_1$ is a defuzzification of the linguistic variable of $a(x)$, and $c_2$ is a defuzzification of the linguistic variable of $a(y)$, respectively. $\sigma_a$ is the standard deviation under the attribute $a$ when the linguistic variables $a(x)$ and $a(y)$ are de-fuzzified.

To handle hybrid attributes, a hybrid distance $(HD)$ is defined based on the different attributes distance as mentioned earlier as follows.

**Definition 16**. Given a $HIS$, for each $x, y \in U$, the hybrid distance $(HD)$ is defined as (39):

$$HD(x, y) = \sqrt{\sum_{k=1}^{m} d^2(a_k(x), a_k(y))} \quad (19)$$

whereas the number of attributes is denoted by $m$, the above content can be summarized as follows (39, 43, 44):

$$d(a(x), a(y)) = \begin{cases} BD(a(x), a(y)), a \text{ is a boolean}, \\ SD(a(x), a(y)), a \text{ is a set-valued}, \\ RD(a(x), a(y)), a \text{ is a real-valued}, \\ LD(a(x), a(y)), a \text{ is a linguistic variable}. \end{cases} \quad (20)$$

## 4. PROPOSED METHOD

In recent years, fuzzy rough set theory has found extensive applications across various fields due to its effectiveness in handling uncertainty and vagueness in data. However, a limitation of the current fuzzy dependency function is its inability to accurately classify subsets of attributes in high-dimensional data. This challenge arises from the intersection operation used to compute fuzzy similarity relations when determining rough approximations, which may not effectively discriminate the membership degrees of these relations in a high-dimensional sample space. As a result, the true relationship between samples may not be accurately represented by the membership function of a fuzzy similarity relation.

### 4. 1. Creating Fuzzy Information Granules

Recently, Hu et al. (45) proposed a combination of the Gaussian kernel function and Euclidean distance to create fuzzy information granules. However, it was noted that the Euclidean distance was inadequate for directly handling hybrid attributes. To address this limitation, Zeng et al. (39) introduced a hybrid distance measure that integrates the Euclidean distance with HD (hybrid distance). This approach enabled the effective construction of fuzzy information granules capable of managing hybrid attributes.

Let $U$ represent a non-empty finite universe, with samples expressed as mmm-dimensional vectors.

$\forall x_i, x_j \in U, x_i = \langle x_{i1}, x_{i2}, \ldots, x_{im} \rangle, x_j = \langle x_{j1}, x_{j2}, \ldots, x_{jm} \rangle.$

The Gaussian kernel function

$$k(x_i, x_j) = exp\ (-\frac{||x_i - x_j||^2}{2\sigma^2}) \quad (21)$$

can be used to compute the similarity between $x_i$ and $x_j$ in wich $||x_i - x_j||$ is the $HD$ distance between $x_i$ and $x_j$. The following relations hold:

(1) $k(x_i, x_j) \in [0,1]$,

(2) $k(x_i, x_j) = k(x_j, x_i)$,

(3) $k(x_i, x_i) = 1$.

and the fuzzy relation $R_G$ computed with Gaussian kernel is given as follows:

$$R_G(x_i, x_j) = exp\ (-\frac{\sum_{k=1}^{m} d^2(a_k(x_i), a_k(x_j))}{2\sigma^2}) \quad (22)$$

**Theorem 1.** According to Sun et al. (21) the Gaussian kernel $k: U \times U \to [0,1]$ satisfies $T_p - transitive$, and the fuzzy relation $R_G$ is a $T_p - equivalence$ relation.

**Proof.**

1) $R_G(x_i, x_i) = 1$ then $Reflexivity$

2) $R_G(x_i, x_j) = R_G(x_j, x_i)$ then $Symmetric$

3) $||x_i - x_j + x_j - x_r||^2 < ||x_i - x_j||^2 + ||x_j - x_r||^2 \Rightarrow$
$-||x_i - x_j||^2 - ||x_j - x_r||^2 < -||x_i - x_r||^2 \Rightarrow$
$exp(\frac{-||x_i - x_j||^2}{2\sigma^2}).\ exp(\frac{-||x_j - x_r||^2}{2\sigma^2}) < exp(\frac{-||x_i - x_r||^2}{2\sigma^2}) \Rightarrow$
$R_G(x_i, x_j).\ R_G(x_j, x_r) < R_G(x_i, x_r) \Rightarrow$
$T_p(R_G(x_i, x_j), R_G(x_j, x_r)) < T_p(R_G(x_i, x_r))$ then $T - transitivity$

### 4. 2. Creating Similarity Relations

As previously mentioned, determining similarity between samples is a significant challenge in fuzzy rough set theory, and addressing this challenge requires the use of distance measures between samples. The decision-making information system employed in this study utilizes a hybrid system that incorporates linguistic variables along with multiple features.

To fully understand the proposed model and accurately compute the similarity distance between samples, let us examine a detailed example that utilizes the formulas discussed earlier.

Based on Table 1, the HD matrix can be calculated. According to the introduced formula, the results hold:

1) $a_1$ is the linguistic variable attribute. Convert linguistic variable into a trapezoidal fuzzy number:

**TABLE 1.** A hybrid information system

| U | Illness rate ($a_1$) | Pain ($a_2$) | Fever temperature ($a_3$) | Syndrome ($a_4$) | D |
|---|---|---|---|---|---|
| $x_1$ | Minor | Yes | 39.2 | $\{C,R,A\}$ | Flu |
| $x_2$ | Minor | Yes | 39.5 | $\{C,R,A\}$ | Flu |
| $x_3$ | Moderate | No | 39.0 | $\{C\}$ | Flu |
| $x_4$ | Minor | Yes | 38.8 | $\{C,R,A\}$ | Rhinitis |
| $x_5$ | Severe | No | 37.3 | $\{R\}$ | Rhinitis |
| $x_6$ | Severe | No | 36.8 | $\{R,A\}$ | Health |
| $x_7$ | Moderate | Yes | 39.0 | $\{C,R,A\}$ | Health |

Minor= (0, 1, 1, 3), Moderate= (2, 4, 6, 6), Severe= (4, 7, 10, 10),

$C_1 = (3a + b + 3c + 2d)/9 = 1.11$, $\quad C_4 = (3a + b + 3c + 2d)/9 = 1.11$,

$LD(a_1(x_1), a_1(x_4)) = \frac{|C_1 - C_4|}{4\sigma_a} = \frac{|1.11-1.11|}{4\times 2.94} = 0$,

2) $a_2$ is a boolean attribute, $BD(a_2(x_1), a_2(x_4)) = 0$,

3) $a_3$ is a real-valued attribute, $RD(a_3(x_1), a_3(x_4)) = \frac{|a_3(x_1) - a_3(x_4)|}{4\sigma_a} = \frac{|39.2-38.8|}{4\times 1.03} = 0.09$,

4) $a_4$ is a set-valued attribute, $SD(a_4(x_1), a_4(x_4)) = 1 - \frac{a_4(x_1) \cap a_4(x_4)}{s} = 1 - \frac{3}{3} = 0$,

And finally,

$$HD(x_1, x_4) = \sqrt{\sum_{k=1}^{4} d^2(a_k(x_1), a_k(x_4))} = 0.09.$$

At the end of this step,an HD matrix is obtained that displays the hybrid distance between the objects in the U set. This matrix is a valuable tool for determining the similarity between the objects in the $U$ set.

$$HD = \begin{bmatrix} 0 & 0.07 & 1.23 & 0.09 & 1.39 & 1.32 & 0.28 \\ & 0 & 1.23 & 0.68 & 1.42 & 1.35 & 0.72 \\ & & 0 & 1.44 & 1.11 & 1.16 & 1.41 \\ & & & 0 & 1.19 & 1.33 & 1.04 \\ & & & & 0 & 0.51 & 1.49 \\ & & & & & 0 & 1.26 \\ & & & & & & 0 \end{bmatrix}$$

In the next step, the $T_P - equivalence$ relation $R_G$, is calculated using the Gaussian kernel function. The fuzzy relation between each two samples can be computed by Equation 22. Let $\sigma^2 = 0.4$. For example:

$$R_G(x_1, x_4) = k(x_1, x_4) = exp(-\frac{HD^2(x_1,x_4)}{2\times\sigma^2}) = exp(-\frac{0.09^2}{2\times 0.4}) = 0.98$$

Therefore,

$$R_G = \begin{bmatrix} 1 & 0.99 & 0.15 & 0.98 & 0.08 & 0.11 & 0.90 \\ & 1 & 0.14 & 0.55 & 0.07 & 0.09 & 0.51 \\ & & 1 & 0.07 & 0.21 & 0.18 & 0.08 \\ & & & 1 & 0.16 & 0.10 & 0.25 \\ & & & & 1 & 0.71 & 0.06 \\ & & & & & 1 & 0.13 \\ & & & & & & 1 \end{bmatrix}$$

Then, the equivalence classes are:

$$\begin{cases} [x_1]_{R_G} = \{\frac{1}{x_1} + \frac{0.99}{x_2} + \frac{0.15}{x_3} + \frac{0.98}{x_4} + \frac{0.08}{x_5} + \frac{0.11}{x_6} + \frac{0.90}{x_7}\} \\ [x_2]_{R_G} = \{\frac{0.99}{x_1} + \frac{1}{x_2} + \frac{0.14}{x_3} + \frac{0.55}{x_4} + \frac{0.07}{x_5} + \frac{0.09}{x_6} + \frac{0.51}{x_7}\} \\ \vdots \\ [x_7]_{R_G} = \{\frac{0.90}{x_1} + \frac{0.51}{x_2} + \frac{0.08}{x_3} + \frac{0.25}{x_4} + \frac{0.06}{x_5} + \frac{0.13}{x_6} + \frac{1}{x_7}\} \end{cases}$$

According to Relation 7, the fuzzy upper approximation of the relation of $R_G$ is defined as follows:

$$\overline{R}_G = \begin{bmatrix} 1 & 0.99 & 0.15 & 0.98 & 0.16 & 0.12 & 0.90 \\ & 1 & 0.15 & 0.97 & 0.9 & 0.11 & 0.89 \\ & & 1 & 0.15 & 0.21 & 0.18 & 0.14 \\ & & & 1 & 0.16 & 0.11 & 0.88 \\ & & & & 1 & 0.71 & 0.09 \\ & & & & & 1 & 0.13 \\ & & & & & & 1 \end{bmatrix}$$

The fuzzy lower approximation of the relation of $R_G$ ($\underline{R}_G$) can be obtained in a similar way, elements of $\underline{R}_G$ indicate certaintly version of $[x]$ under fuzzy relation $R$ and elements of $\overline{R}_G$ indicate possiblity version of $[x]$ under fuzzy relation $R$. The important point is that $\underline{R}_G$ is not a fuzzy equivalence relation.

Any of the above relations ($R_G, \overline{R}_G$) can be used to create a new model based on the opinion of the decision-maker. The state of model is called normal when $R_G$ is used for modeling and state is called optimistic when $\overline{R}_G$ is used for modeling.

**4. 3. Creating a Set of Acceptable Indexes of the Problem** Since $T_p$ is a special $T-norm$, according to Theorem 1, it can be used $R_G$ as a similarity relation. According to the obtained similarity equivalence relation and considering the classification of objects, Table 2 can be drawn as follows:

In Table 2, the similarity of objects can be seen by considering their classification. The final goal of this research is to select the features, and it is intended to select the features in such a way that the classification of the objects does not deteriorate, and calculations are applied in such a way that fewer calculations are done in the final model presented. Therefore, for making a model, the objects that are in the same class will not be considered and a constant number is considered as the degree of similarity criterion, which is displayed with $\delta$ $(0 \le \delta \le 1)$.

The choice of delta size depends on the opinion of the decision-maker, according to the $R_G$ relation and delta size, two sets $G_1$ and $G_2$ which are given below are made: Let $U$ be a non-empty finite universe and divided into r crisp equivalence classes by $D$, $D = \{d\}$ is the decision attributes, $U/D = \{D_1, D_2, \ldots, D_r\}$.
If $(x_i, x_j) \in D_l \times D_m (l \neq m)$ then,

$$G_1 = \{(x_i, x_j) | R(x_i, x_j) \le \delta\},$$
$$G_2 = \{(x_i, x_j) | R(x_i, x_j) > \delta\} \tag{23}$$

For example, according to Table 2,
$D_1 = \{x_1, x_2, x_3\}, D_2 = \{x_4, x_5\}, D_3 = \{x_6, x_7\}$.
if $\delta = 0.85$, then

$$G_1 = \{(x_1, x_5), (x_1, x_6), (x_2, x_4), (x_2, x_5), (x_2, x_6), (x_2, x_7), (x_3, x_4), (x_3, x_5), (x_3, x_6), (x_3, x_7), (x_4, x_6), (x_4, x_7), (x_5, x_6), (x_5, x_7)\}.$$

$$G_2 = \{(x_1, x_4), (x_1, x_7)\}.$$

Considering that the set of $G_2$ represents the objects that belong to the fuzzy rough set and the similarity of the features becomes fixed or more by feature selection, $G_2$ is to be ignored.

To describe the proposed new method, the set $S_\delta$ is defined:

$$S_\delta = \{(i,j) | (x_i, x_j) \in G_1\} \tag{24}$$

**4. 4. Creating Constraint and Objective Function of the Problem** By reducing each feature, the distance between objects can be decreased or maintained at the same level, and this can be utilized as an effective method for measuring the similarity between objects. The greater the similarity between the objects, the smaller the distance value. Therefore, the similarity relation between objects can be increased or maintained at the same level by reducing each feature.

This is because each feature represents a certain aspect or characteristic of the objects being analyzed, and features that do not contribute significantly to the similarity between objects can be removed. By focusing on the most informative and relevant features, we can better understand the underlying patterns and relations in the data, and more accurate and efficient models can be developed.

By assuming a feature reduction and removing one or several features, the objects are in the same classes, and the relations whose value is greater than the measure of the degree of similarity $\delta$ are ignored. But other relations would still have to be inferior to the criterion of the degree of similarity.

This article is presented as a constraint in problem modeling as below:

$$e^{\left(-\frac{\chi_k HD^2(x_i,x_j)}{2\sigma^2}\right)} \le \delta \rightarrow -\frac{\chi_k HD^2(x_i,x_j)}{2\sigma^2} \le \ln\delta, (i,j) \in S_\delta \tag{25}$$

**TABLE 2.** The similarity between objects with considering their classes

| | | **1** | | | **2** | | **3** | |
|---|---|---|---|---|---|---|---|---|
| **Class** | **Object** | $x_1$ | $x_2$ | $x_3$ | $x_4$ | $x_5$ | $x_6$ | $x_7$ |
| 1 | $x_1$ | 1 | 0.99 | 0.15 | 0.98 | 0.08 | 0.11 | 0.90 |
| | $x_2$ | | 1 | 0.14 | 0.55 | 0.07 | 0.09 | 0.51 |
| | $x_3$ | | | 1 | 0.07 | 0.21 | 0.18 | 0.08 |
| 2 | $x_4$ | | | | 1 | 0.16 | 0.10 | 0.25 |
| | $x_5$ | | | | | 1 | 0.71 | 0.06 |
| 3 | $x_6$ | | | | | | 1 | 0.13 |
| | $x_7$ | | | | | | | 1 |

Let $\chi_k \in \{0,1\}$,$\chi_k = 1$ means that the feature is selected and $\chi_k = 0$ means that the feature is not selected. $k = \{1,2,\dots,p\}$, and $p$ represents the number of attributes. The goal of this problem is to reduce features, so the objective function will be formulate as follows:

$$Min \sum_{k=1}^{p} \chi_k \quad (26)$$

**4. 5. Obtaining Optimization Problem** Based on the information provided above, the problem will be modeled as follows:

The set of objects and their corresponding features will be defined, and a hybrid distance measure will be employed to determine the relationships between these objects. Additionally, excluding objects that belong to the same classes, as well as relationships with values greater than the degree of similarity, can simplify the model and help prevent overfitting.

Subsequently, the constraints and objective function of the problem will be generated, and a novel model will be defined as follows:

$$Min \sum_{k=1}^{p} \chi_k$$

$$s.t. -\frac{\sum_{k=1}^{p} \chi_k \ d^2(a_k(x_i), a_k(x_j))}{2\sigma^2} \le \ln\delta \quad (27)$$

$$(i,j) \in S_\delta, \chi_k \in \{0,1\}$$

The problem can now be addressed by selecting an appropriate meta-heuristic algorithm to find an approximate solution to the proposed model. In the following sections, we will apply the black hole (BH) algorithm, which has been assessed as suitable for such problems.

## 5. EXPERIMENTAL EVALUATION

In this research, we will validate the effectiveness of the proposed model using public datasets. To achieve this, datasets will be downloaded from the UC-Irvine Machine Learning Repository, and classification algorithms will be applied to these datasets after employing various feature selection algorithms to evaluate the model's effectiveness.

Three types of experimental results will be presented:

1. The feature selection results of FSbuHD in two states: normal and optimistic.
2. A comparison of the previously tested algorithms with FSbuHD.
3. A comparison of accuracy, precision, recall, and the Matthews correlation coefficient.

Detailed explanations of these results are provided below.

**5. 1. Metaheuristic Algorithms** The goal of feature selection is to identify the smallest possible number of attributes that can achieve the same or similar classification accuracy as the entire set of attributes. While there may be several possible subsets of attributes that can accomplish this, it is sufficient to find just one of them. The black hole optimization algorithm is a type of meta-heuristic algorithm inspired by nature. It simulates the phenomenon of black holes, which are formed from massive stars with strong gravitational forces. The algorithm begins with a specific population size of potential solutions, which are then evaluated. The best solution is selected as the "black hole" (46, 47).

A total of eight datasets were utilized from the UCI machine learning repository to assess the effectiveness of the model. These datasets comprise a variety of features including single and mixed types, and also consist of both binary and multiple classes. A summary of fundamental descriptive characteristics for each of the datasets used is provided in Table 3.

Evaluation of feature selection using different algorithms (FARNem, WARA, CfsSubsetEval, and RSFSAID) is shown in Table 4.

FARNeM is used, $\delta$, to determine the neighborhood set of objects, WARA is used a parameter, $\epsilon$, which serves as a threshold for feature selection, CfsSubsetEval, can be implemented without any parameters, and RSFSAID, is used a parameter, $\delta$, which is determined in the same way as in the FARNeM. For the experiments, the values of $\delta$ and $\epsilon$ were fixed to increase from 0 to 0.99 in increments of 0.01. The best results were registered (48).

The results of the proposed model using $\delta$ and $\sigma = 0.2$ in different data sets in two normal and optimistic states are shown in the last two columns of Table 4 for comparison with other algorithms. The value of $\delta$ is considered from 0 to 0.9 incrementally, and with step 0.1, the best results were registered.

In Table 5, the set of selected features in different datasets that were chosen by the FSbuHD method can be seen.

**5. 2. Result of Feature Selection** Tables 4 and 5 present the results of FSbuHD in both normal and optimistic states. The evaluation of feature selection using FARNeM, WARA, CfsSubsetEval, RSFSAID, and FSbuHD for different datasets is detailed in Table 4.

The results displayed in Table 4 indicate that FSbuHD is a highly effective method for selecting a subset of features. It is important to note that the features selected by different algorithms may vary, as the feature evaluation procedures and selection algorithms employed can differ. However, the efficacy of any feature selection algorithm can only be verified through the use of classification algorithms. By evaluating the performance of the selected features with classification algorithms, we can assess the effectiveness of the feature selection algorithm in question.

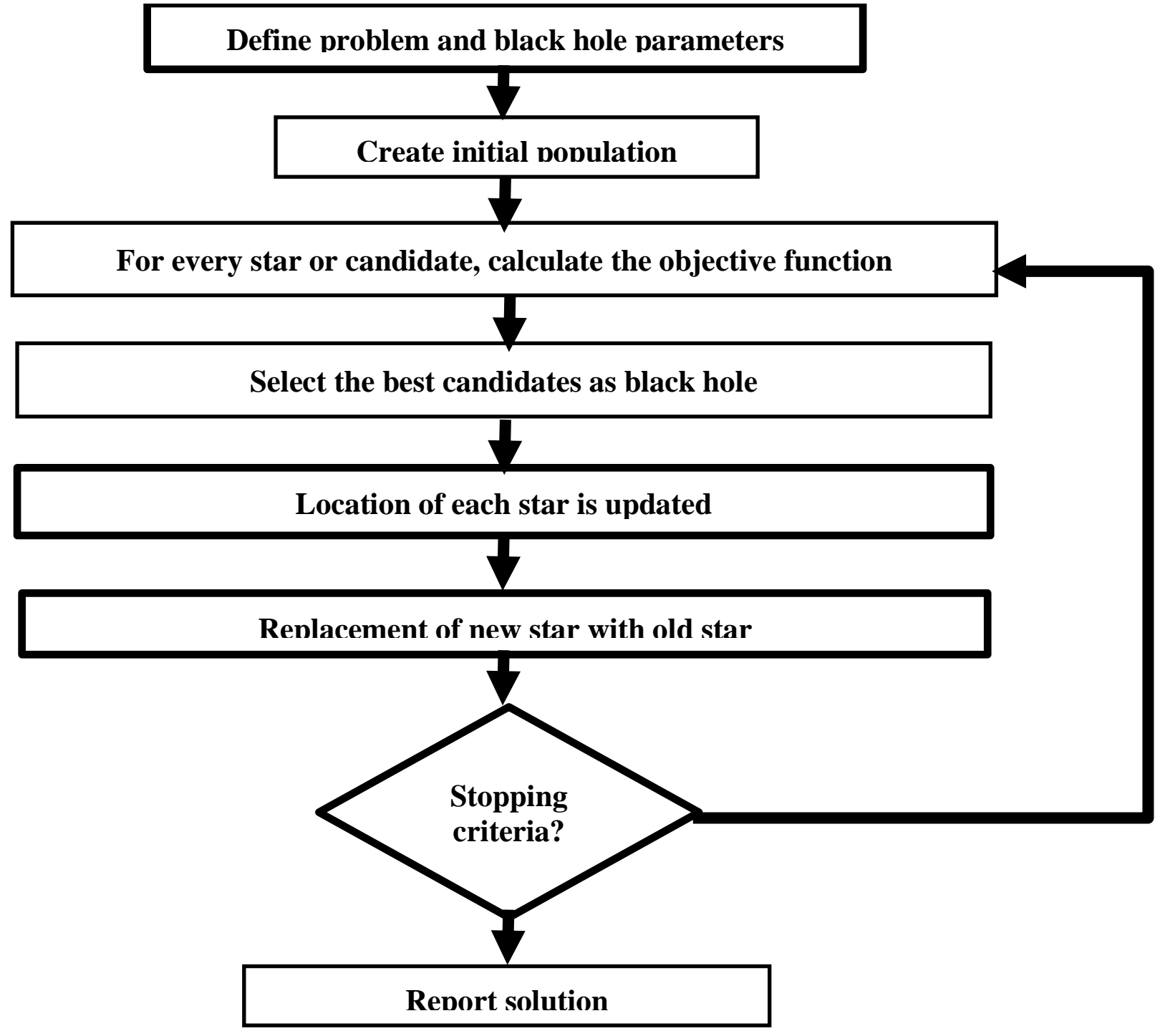

**Figure 2.** Framework of feature selection

**TABLE 3.** Summary of the datasets

| NO. | Name | # I | # F | # C | Data Type |
|---|---|---|---|---|---|
| 1 | crx | 690 | 15 | 2 | Hybrid |
| 2 | australian | 690 | 14 | 2 | Hybrid |
| 3 | heart | 270 | 13 | 2 | Hybrid |
| 4 | ionosphere | 351 | 34 | 2 | Hybrid |
| 5 | segment | 2308 | 19 | 7 | Numerical |
| 6 | wpbc | 198 | 33 | 2 | Numerical |
| 7 | zoo-3 | 101 | 16 | 7 | Numerical |
| 8 | wdbc | 569 | 30 | 2 | Numerical |

#I, denotes the number of instances. #F, denotes the number of features. #C, denotes the number of classes.

**TABLE 4.** Feature selection using different algorithms

| NO. | Dataset | FARNem-F (#) | WARA-F (#) | CfsSubset Eval-F (#) | RSFSAID -F (#) | FSbuHD-F(#) | |
|---|---|---|---|---|---|---|---|
| | | | | | | Normal state | Optimistic state |
| 1 | crx | 14 | 13 | 7 | 9 | 6 | 6 |
| 2 | australian | 14 | 12 | 7 | 6 | 6 | 4 |
| 3 | heart | 11 | 11 | 8 | 5 | 7 | 6 |
| 4 | ionosphere | 16 | 8 | 17 | 8 | 15 | 14 |
| 5 | segment | 10 | 11 | 3 | 3 | 10 | 9 |
| 6 | wpbc | 12 | 1 | 13 | 12 | 8 | 5 |
| 7 | zoo-3 | 5 | 5 | 7 | 6 | 5 | 4 |
| 8 | wdbc | 8 | 5 | 11 | 8 | 13 | 13 |

**TABLE 5.** Set of feature selection using FSbuHD

| NO. | Dataset | State | FSbuHD-F(#) | FSbuHD-F(S) |
|---|---|---|---|---|
| 1 | crx | Normal | 6 | [4,9,10,11,12,15] |
| | | Optimistic | 6 | [2,5,9,10,11,14] |
| 2 | australian | Normal | 6 | [4,5,8,10,12,14] |
| | | Optimistic | 4 | [5,8,12,14] |
| 3 | heart | Normal | 7 | [1,2,7,8,9,12,13] |
| | | Optimistic | 6 | [2,3,7,9,12,13] |
| 4 | ionosphere | Normal | 15 | [2,3,5,6,7,8,11,18,20,21,22,23,24,29,34] |
| | | Optimistic | 14 | [1,2,3,5,7,11,13,16,18,19,21,22,33,34] |
| 5 | segment | Normal | 10 | [1,2,3,5,8,11,14,16,18,19] |
| | | Optimistic | 9 | [2,3,13,14,15,16,17,18,19] |
| 6 | wpbc | Normal | 8 | [1,2,8,21,23,24,26,31] |
| | | Optimistic | 5 | [2,7,13,17,20] |
| 7 | zoo-3 | Normal | 5 | [2,3,4,6,9] |
| | | Optimistic | 4 | [4,6,9,10] |
| 8 | wdbc | Normal | 13 | [1,2,3,7,8,10,14,15,18,23,24,27,28] |
| | | Optimistic | 13 | [1,2,3,7,8,13,14,15,18,23,25,26,28] |

F(S), denotes the set of features that are selected.

Therefore, it is imperative to conduct a thorough analysis of the selected features using various classification techniques to ensure that the proposed feature selection approach is both accurate and reliable. Overall, the results presented in Table 4 suggest that FSbuHD is a promising method for feature selection and merits further investigation.

### 5. 3. Evaluation by Using the Classification Performance

In this research, three different algorithms—namely, Linear SVM, KNN, and Complex Tree—have been employed to evaluate the effectiveness of the proposed model for feature selection. The results of feature selection across different algorithms and various datasets have been analyzed and compared.

Linear SVM is utilized for linearly separable data, while K-nearest neighbors (KNN) is a non-parametric, supervised learning classifier that makes classifications or predictions based on the proximity to other data points (with k=3 considered in this experiment). The Complex Tree is a hierarchical decision support model that employs decision trees to represent an algorithm containing conditional control statements, chance event outcomes, resource costs, and utility.

Experiments were conducted using a five-fold cross-validation framework. Each dataset was divided into five sections, with one section designated as the testing set and the other four used for training. Each section is rotated as the testing set, resulting in five experiments for each dataset. For each experiment, the training set undergoes preprocessing using feature selection algorithms. A classifier is then constructed based on the training set after feature selection. Finally, the performance of the classifier is evaluated by applying it to the testing set.

In a binary classification problem, the classification results are labeled as positive (P) and negative (N), leading to the creation of a confusion matrix based on these labels. The confusion matrix is a standardized table that compares the number of instances in different groups according to binary classification.

| Confusion Matrix | | | |
|---|---|---|---|
| | | Prediction | |
| Actual | Positive | TP | FN |
| Class | Negative | FP | TN |

The accuracy, precision, recall, and Matthews correlation coefficient (MCC) are used to evaluate different algorithms and determine which one is most effective in comparison to others. All the metrics mentioned above are derived from the confusion matrix. This section analyzes the accuracy, precision, recall, and MCC for FARNeM, WARA, CfsSubsetEval, RSFSAID, and FSbuHD (normal state).

From the eight datasets presented in Table 3, three datasets are selected at random for evaluation, and the results are obtained for these subsets. It is possible to calculate the performance metrics for all the datasets listed in the table, although some calculations may have been omitted for brevity. The Classification Learner app in MATLAB R2017a is utilized to prepare Tables 6, 7, 8, and 9.

## 5. 4. Performance Measurement

1) Accuracy:
The percentage of events that were successfully predicted is compared with all the predictions.

$$ACC = \frac{TP+TN}{TP+TN+FP+FN} \tag{28}$$

2) Precision:
The ratio of all true positives to all positive predictions is presented as follows,

$$Precision = \frac{TP}{TP+FP} \tag{29}$$

3) Recall:
The division of true positives by positive results indicates how many potential positives were found by the model (49).

$$Recall = \frac{TP}{TP+FN} \tag{30}$$

4) Matthew correlation coefficient:
Various indicators have been introduced to evaluate the classification, one of them is Matthew correlation coefficient (MMC) index. MMC produces a more informative and truthful score in evaluating binary classification than accuracy and F1 score (50, 51), which is calculated as follows;

$$MCC = \frac{TP\times TN-FP\times FN}{\sqrt{(TP+FP)(TP+FN)(TN+FP)(TN+FN)}} \tag{31}$$

Four tables and figures will be presented in the following, drawn based on each of them (Tables 6 to 9 and Figures 3 to 14). The comparison of accuracy (Table 6, Figures 3, 4 and 5), precision (Tables 7, Figures 6, 7 and 8), recall (Table 8, Figures 9, 10 and11), MMC (Table 9, Figures 12, 13 and 14) according to three different data sets and the UCI repository (52) based on three classification algorithms will be shown.

**TABLE 6.** Compare accuracy

| Dataset | Result of train | Original | FARNem-F(#) | WARA-F(#) | CfsSubset Eval-F(#) | RSFSAID-F(#) | FSbuHD-F(#) |
|---|---|---|---|---|---|---|---|
| | Linear SVM | 0.7626 | 0.7828 | 0.7626 | 0.7677 | 0.7626 | 0.7778 |
| wpbc | KNN | 0.7121 | 0.7374 | 0.6717 | 0.7071 | 0.6919 | 0.7223 |
| | Complex Tree | 0.6869 | 0.7020 | 0.6313 | 0.7222 | 0.6869 | 0.7222 |
| | Linear SVM | 0.8481 | 0.8370 | 0.8407 | 0.8556 | 0.8370 | 0.8370 |
| heart | KNN | 0.7852 | 0.7667 | 0.8074 | 0.8111 | 0.8037 | 0.8111 |
| | Complex Tree | 0.7556 | 0.7556 | 0.7741 | 0.7963 | 0.7815 | 0.7963 |
| | Linear SVM | 0.8536 | 0.8536 | 0.8522 | 0.8551 | 0.8551 | 0.8522 |
| australian | KNN | 0.8290 | 0.8290 | 0.8319 | 0.8609 | 0.8304 | 0.8435 |
| | Complex Tree | 0.8304 | 0.8304 | 0.8348 | 0.8275 | 0.8319 | 0.8435 |

The values for different algorithms are underlined if greater than the original, and the greatest value is bold.

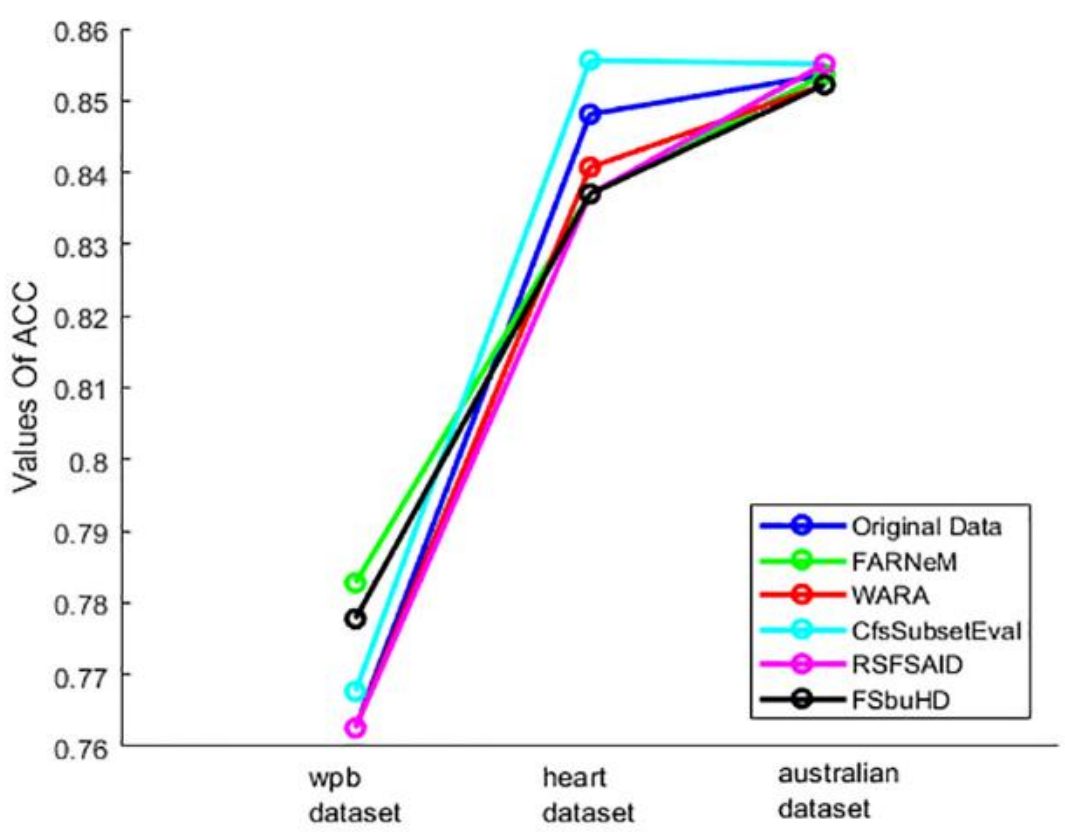

**Figure 3.** The values of ACC when using SVM classifier

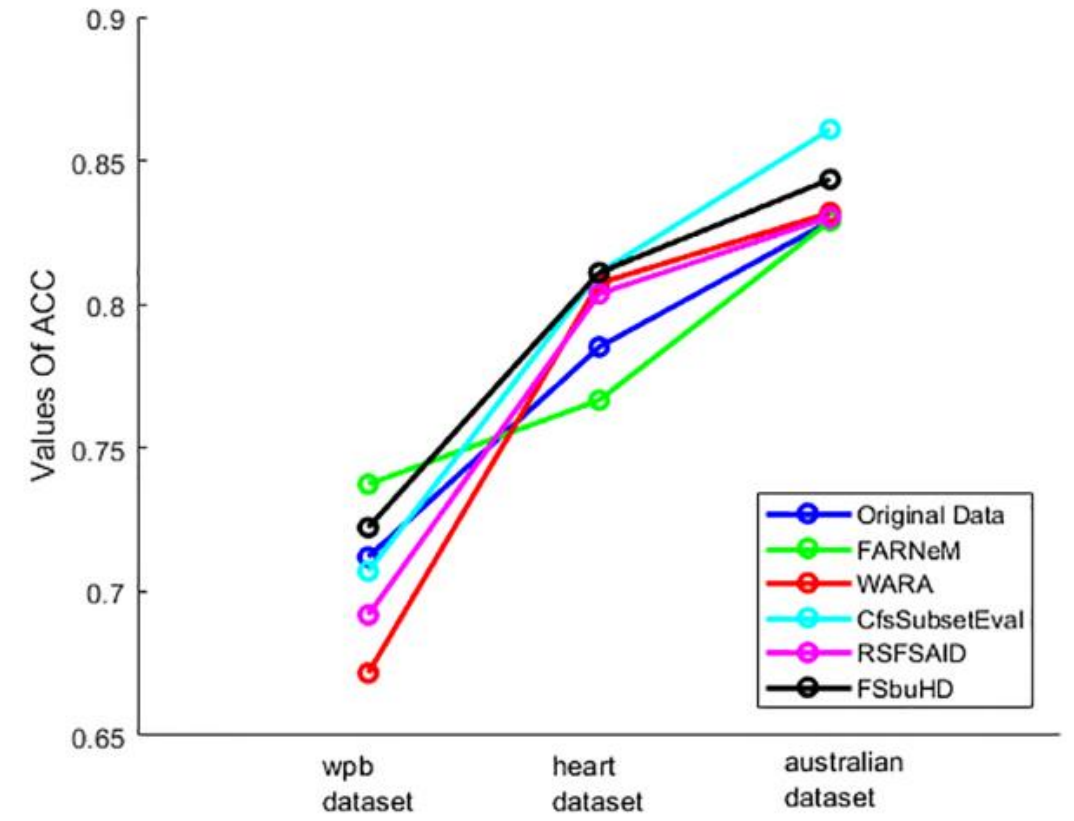

**Figure 4.** The values of ACC when using KNN classifier

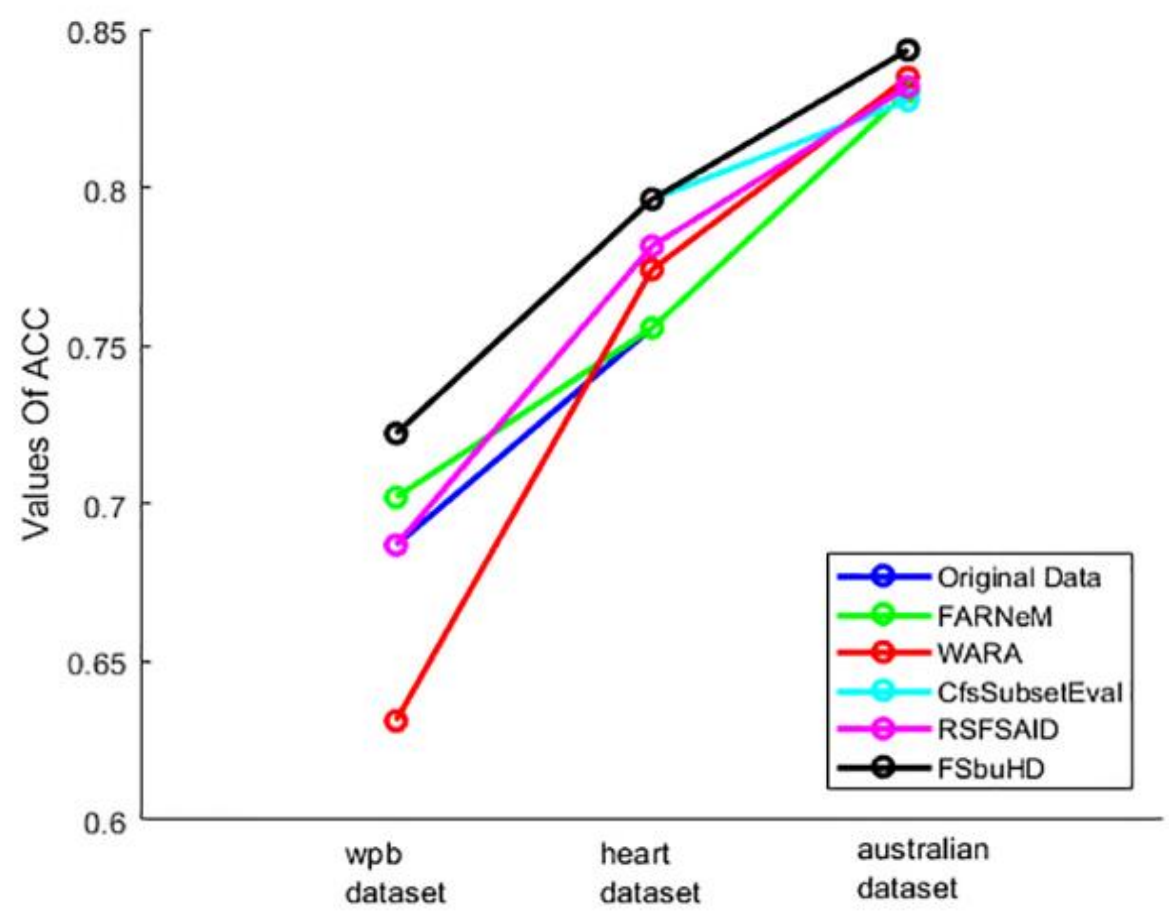

**Figure 5.** The values of ACC when using Complex Tree classifier

**TABLE 7.** Compare precision

| Dataset | Result of train | Original | FARNem-F(#) | WARA-F(#) | CfsSubset Eval-F(#) | RSFSAID-F(#) | FSbuHD-F(#) |
|---|---|---|---|---|---|---|---|
| | Linear SVM | 0.5000 | **0.6111** | _ | 0.5185 | _ | 0.5882 |
| wpbc | KNN | 0.3750 | **0.4242** | 0.2188 | 0.3429 | 0.3056 | 0.3939 |
| | Complex Tree | 0.3404 | 0.3636 | 0.1750 | **0.4000** | 0.3256 | **0.4000** |
| | Linear SVM | 0.8496 | 0.8393 | 0.8348 | **0.8649** | 0.8167 | 0.8455 |
| heart | KNN | 0.7500 | 0.7395 | 0.7881 | 0.7899 | **0.7913** | 0.7899 |
| | Complex Tree | 0.7213 | 0.7213 | 0.7438 | **0.7982** | 0.7607 | 0.7600 |
| | Linear SVM | 0.7845 | 0.7845 | 0.7839 | **0.7867** | **0.7867** | 0.7839 |
| australian | KNN | 0.8161 | 0.8161 | 0.8013 | **0.8328** | 0.8105 | 0.8241 |
| | Complex Tree | 0.8167 | 0.8167 | **0.8185** | 0.8176 | 0.8032 | 0.7867 |

The values for different algorithms are underlined if greater than the original, and the greatest value is bold.

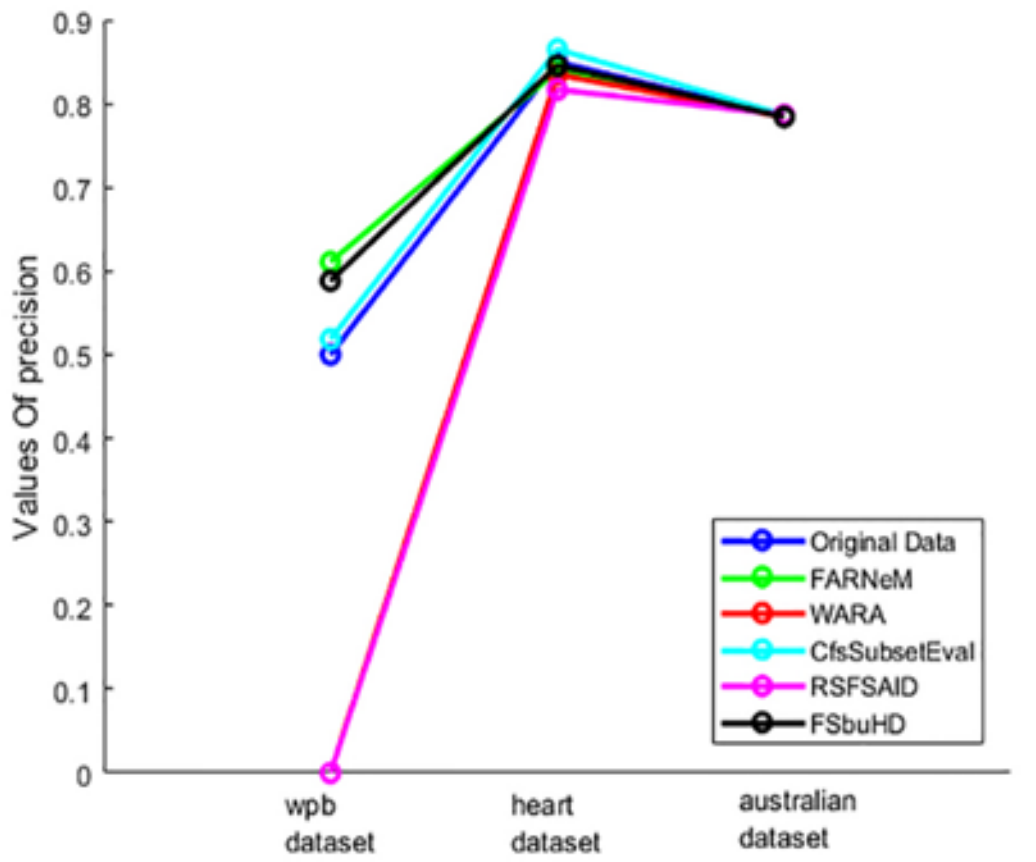

**Figure 6.** The values of precision when using SVM classifier

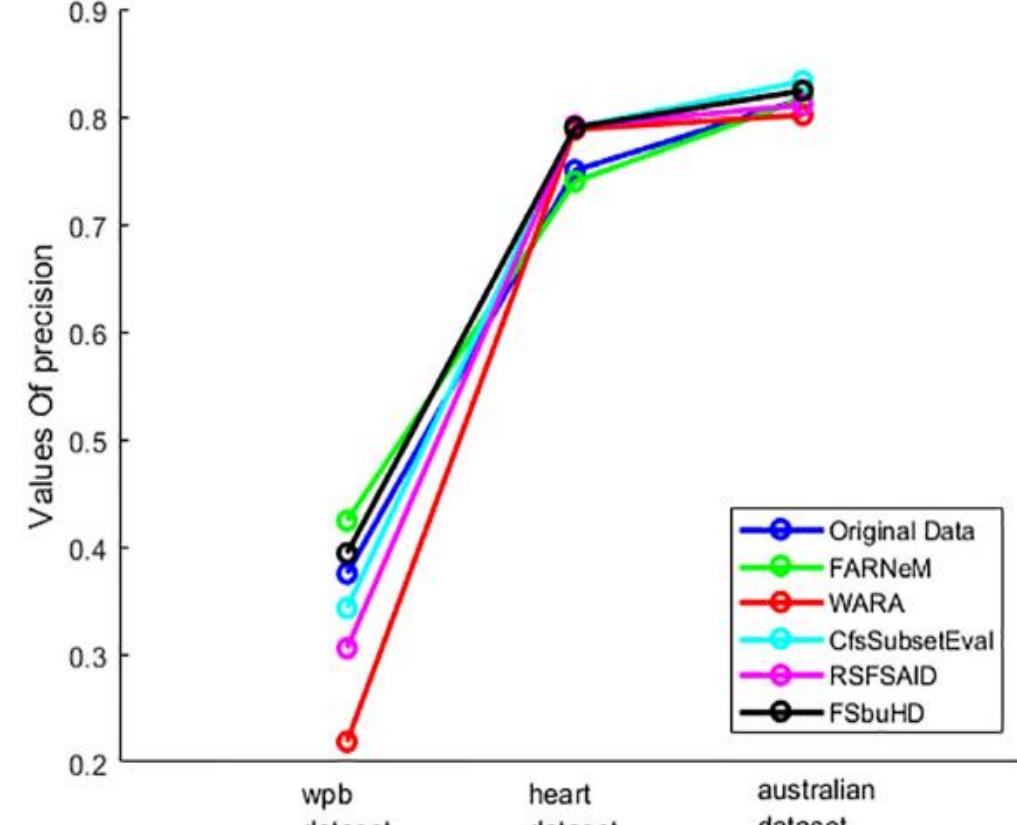

**Figure 7.** The values of precision when using KNN classifier

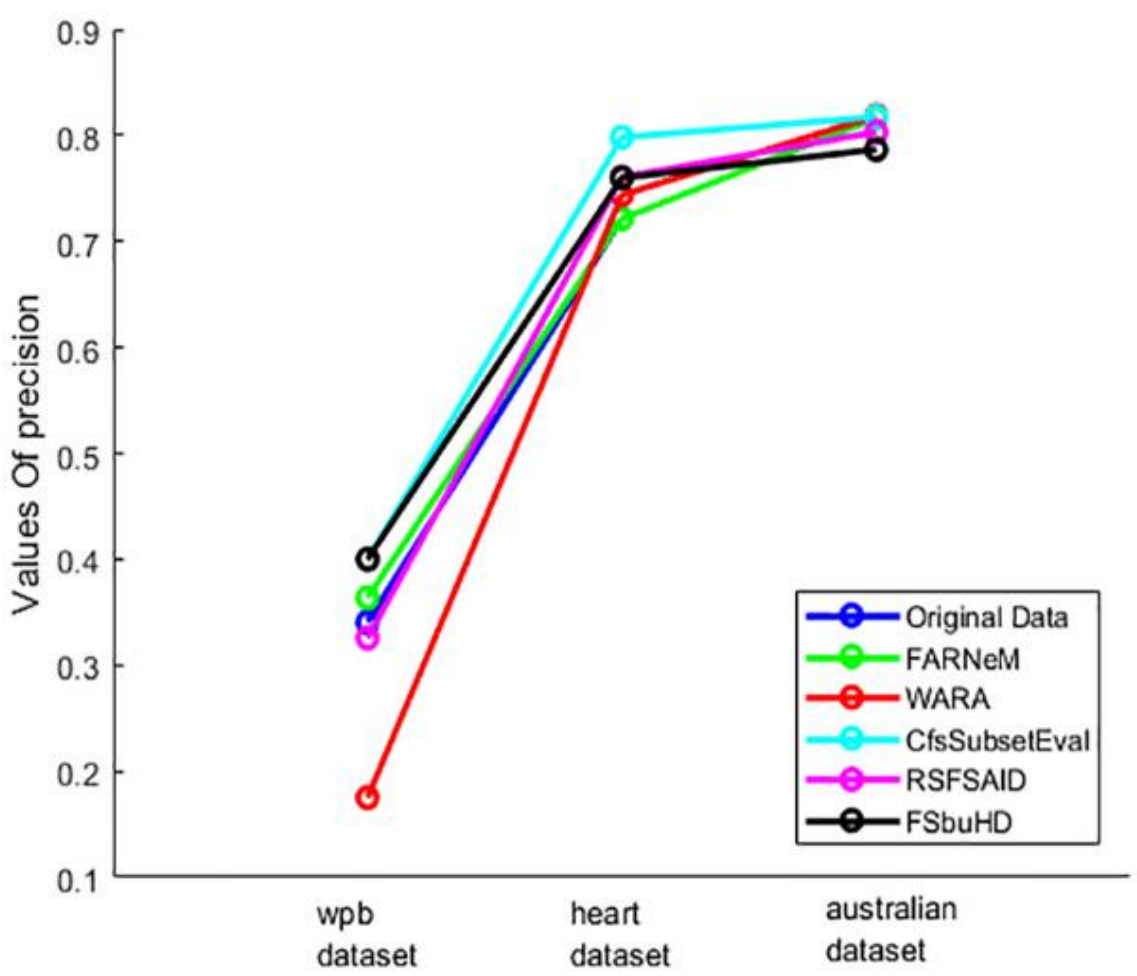

**Figure 8.** The values of precision when using Complex Treeclassifier

**TABLE 8.** Compare recall

| Dataset | Result of train | Original | FARNem-F(#) | WARA-F(#) | CfsSubsetEval-F(#) | RSFSAID-F(#) | FSbuHD-F(#) |
|---|---|---|---|---|---|---|---|
| | Linear SVM | 0.1702 | 0.2340 | 0 | 0.2979 | 0 | 0.2128 |
| wpbc | KNN | 0.3191 | 0.2979 | 0.1489 | 0.2553 | 0.2340 | 0.2766 |
| | Complex Tree | 0.3404 | 0.3404 | 0.1489 | 0.3404 | 0.2979 | 0.3404 |
| | Linear SVM | 0.8000 | 0.7833 | 0.8000 | 0.8000 | 0.8167 | 0.7750 |
| heart | KNN | 0.7750 | 0.7333 | 0.7750 | 0.7833 | 0.7583 | 0.7833 |
| | Complex Tree | 0.7333 | 0.7333 | 0.7500 | 0.7250 | 0.7417 | 0.7917 |
| | Linear SVM | 0.9251 | 0.9251 | 0.9218 | 0.9251 | 0.9251 | 0.9218 |
| australian | KNN | 0.7948 | 0.7948 | 0.8274 | 0.8599 | 0.8078 | 0.8241 |
| | Complex Tree | 0.7980 | 0.7980 | 0.8078 | 0.7883 | 0.8241 | 0.8893 |

The values for different algorithms are underlined if greater than the original, and the greatest value is bold.

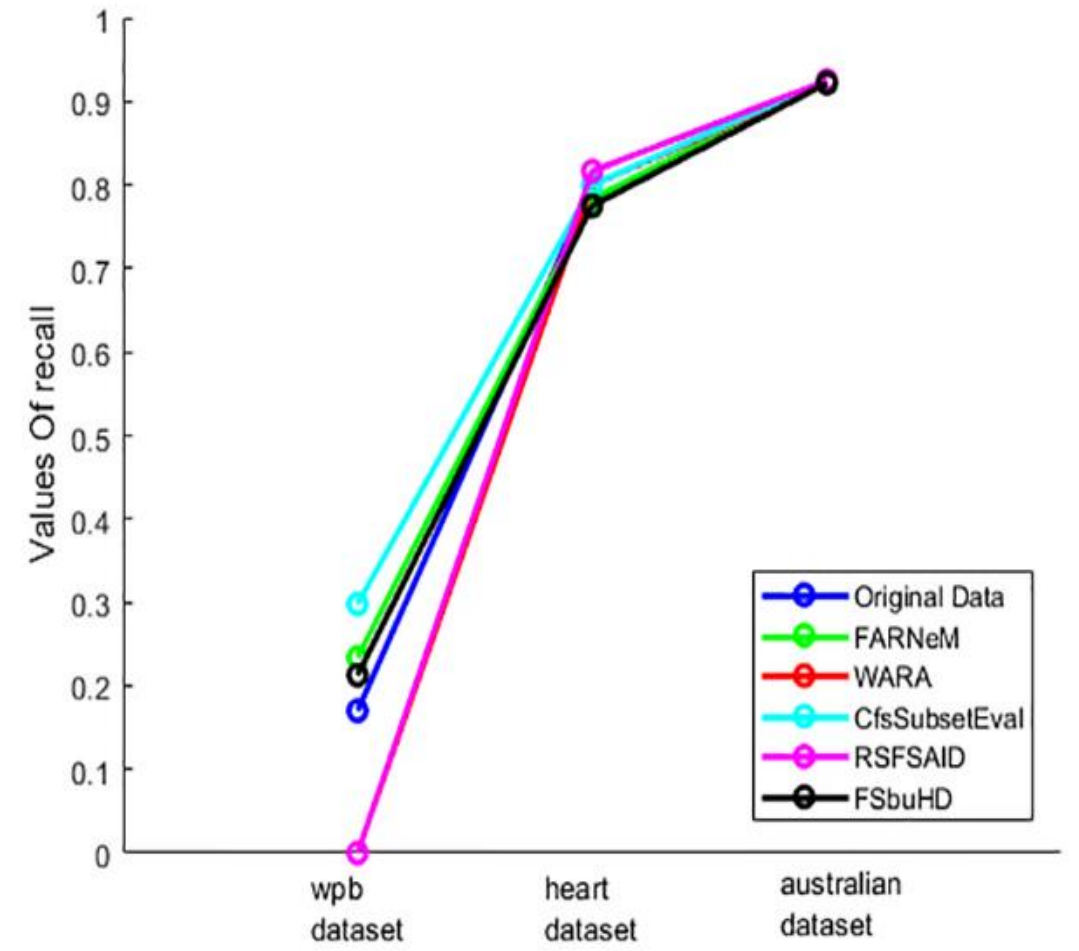

**Figure 9.** The values of recall when using SVM classifier

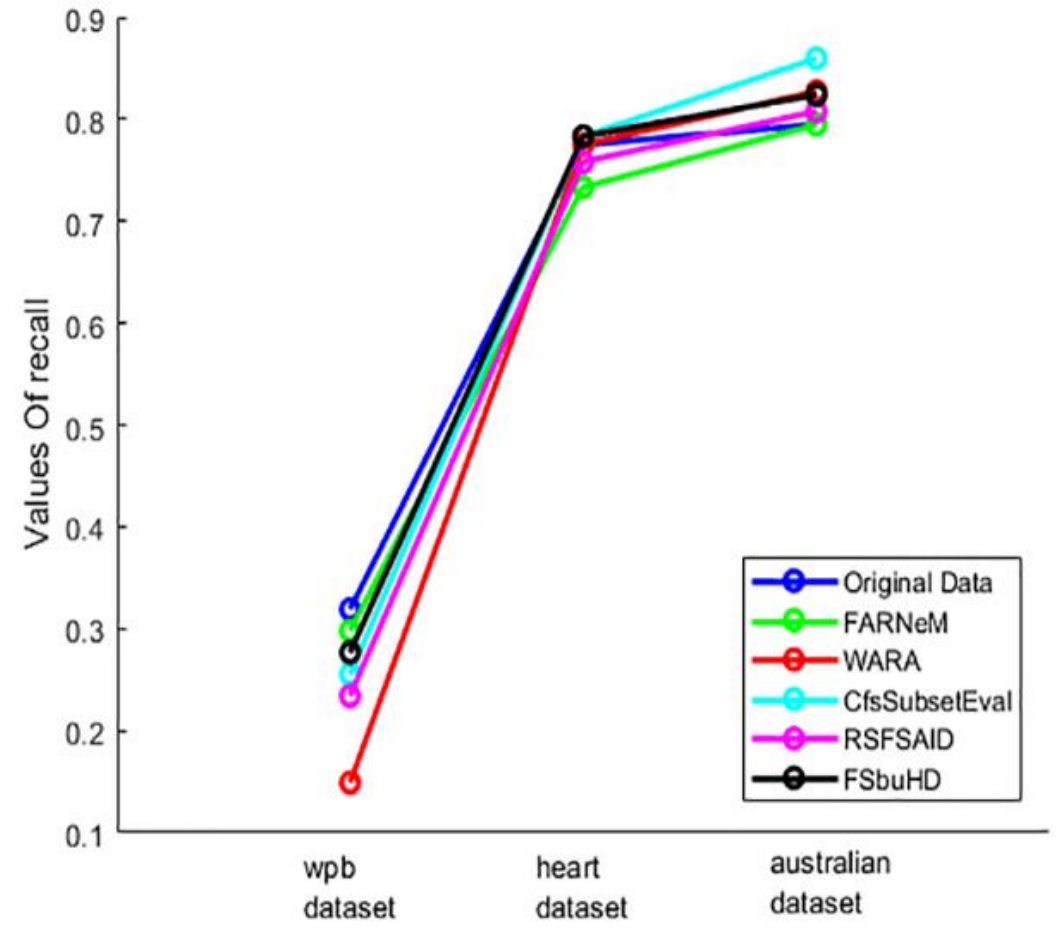

**Figure 10.** The values of recall when using KNN classifier

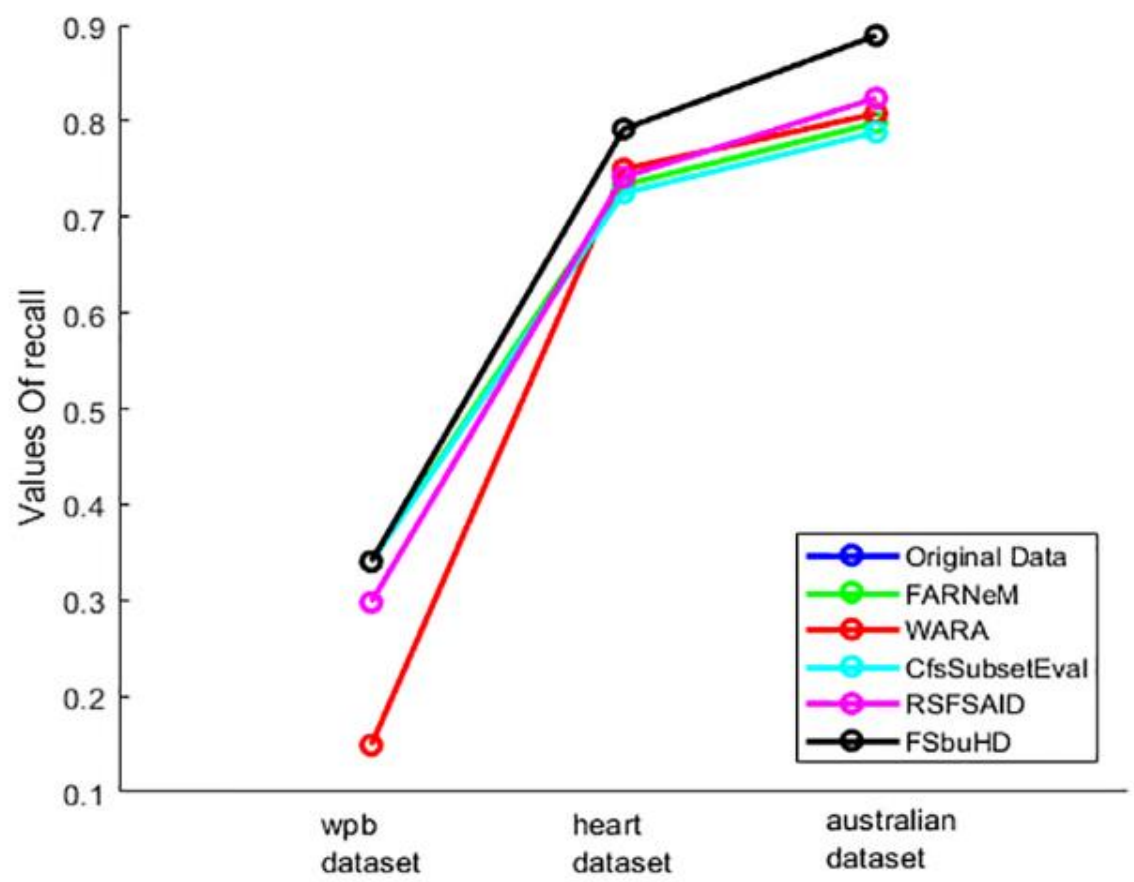

**Figure 11.** The values of recall when using Complex Tree classifier

**TABLE 9.** Compare MMC

| Dataset | Result of train | Original | FARNem-F(#) | WARA-F(#) | CfsSubsetEval-F(#) | RSFSAID-F(#) | FSbuHD-F(#) |
|---|---|---|---|---|---|---|---|
| | Linear SVM | 0.1830 | **0.2778** | - | 0.2626 | - | 0.2527 |
| wpbc | KNN | 0.1628 | **0.1964** | −0.0192 | 0.1149 | 0.0755 | 0.1646 |
| | Complex Tree | 0.1351 | 0.1586 | −0.0738 | **0.1923** | 0.1092 | **0.1923** |
| | Linear SVM | 0.6917 | 0.6690 | 0.6766 | **0.7069** | 0.6700 | 0.6691 |
| heart | KNN | 0.5667 | 0.5271 | 0.6094 | **0.6172** | 0.6013 | **0.6172** |
| | Complex Tree | 0.5059 | 0.5059 | 0.5429 | 0.5857 | 0.5565 | **0.5896** |
| | Linear SVM | 0.7179 | 0.7179 | 0.7146 | **0.7204** | **0.7204** | 0.7146 |
| australian | KNN | 0.6531 | 0.6531 | 0.6610 | **0.7195** | 0.6566 | 0.6831 |
| | Complex Tree | 0.6561 | 0.6561 | 0.6651 | 0.6499 | 0.6607 | **0.6918** |

The values for different algorithms are underlined if greater than the original, and the greatest value is bold.

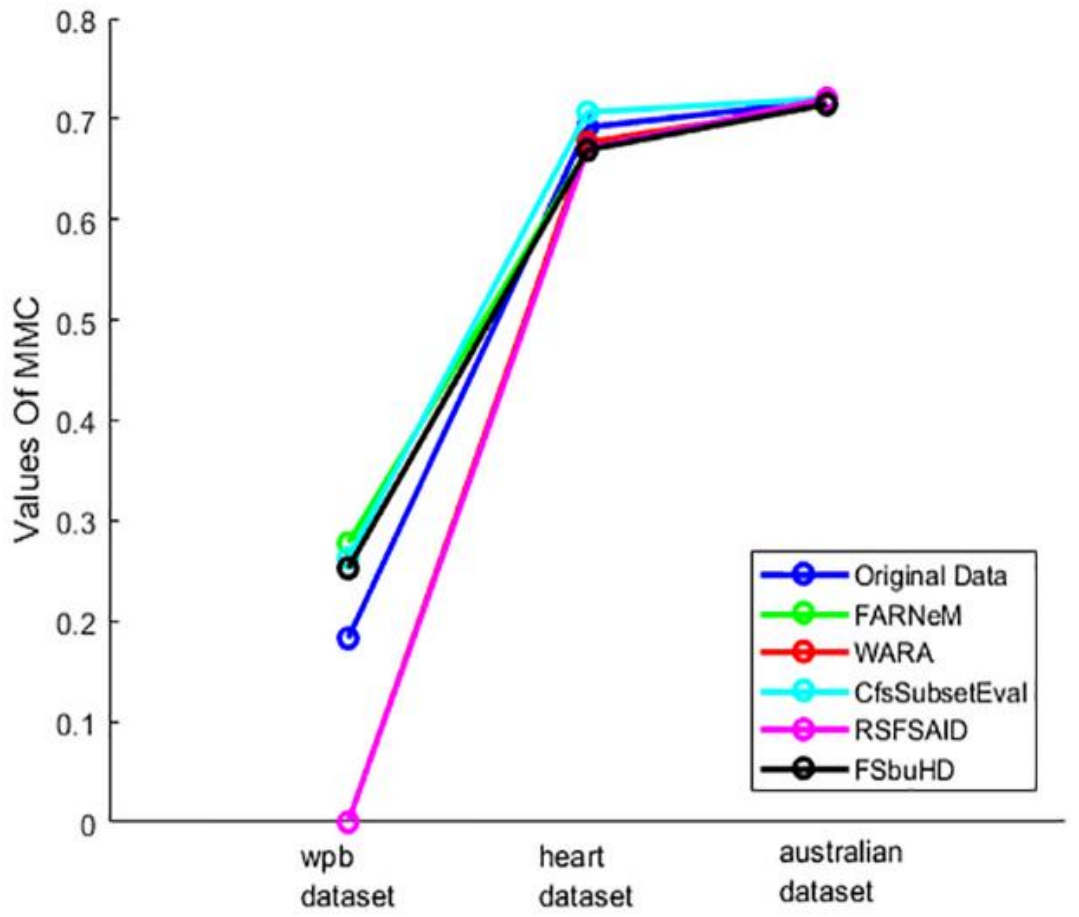

**Figure 12.** The values of MMC when using SVM classifier

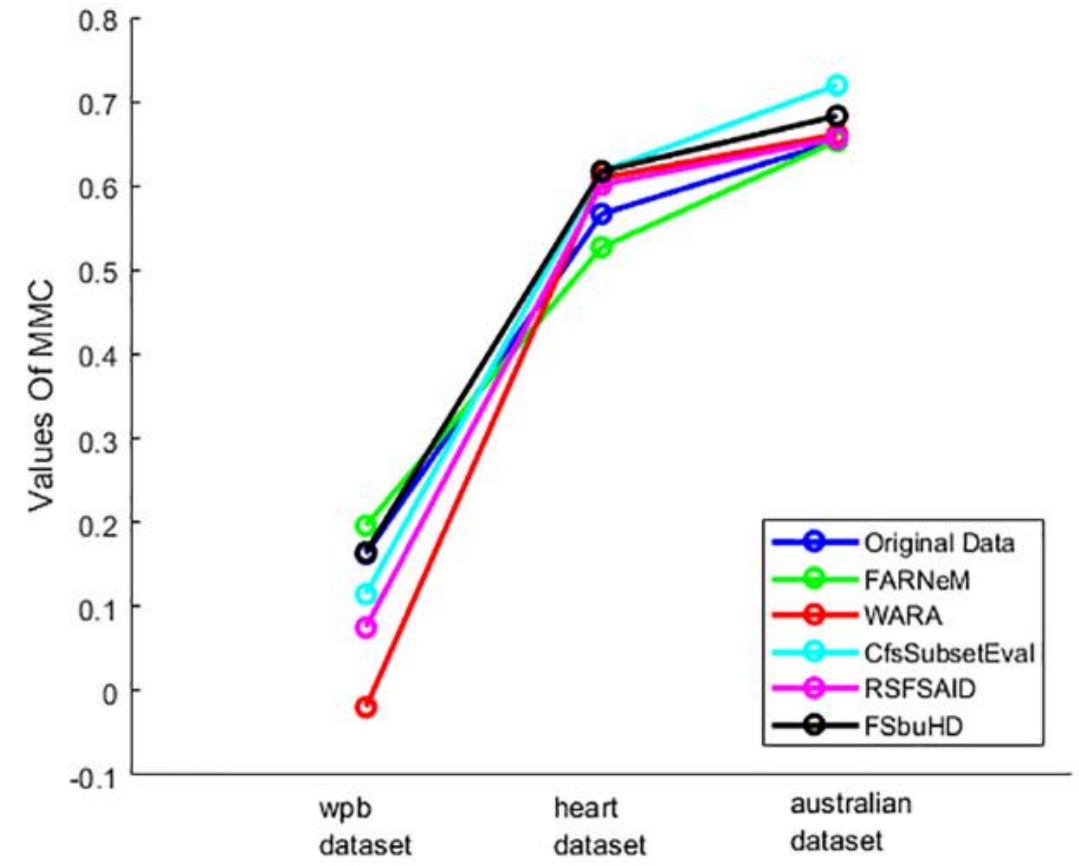

**Figure 13.** The values of MMC when using KNN classifier

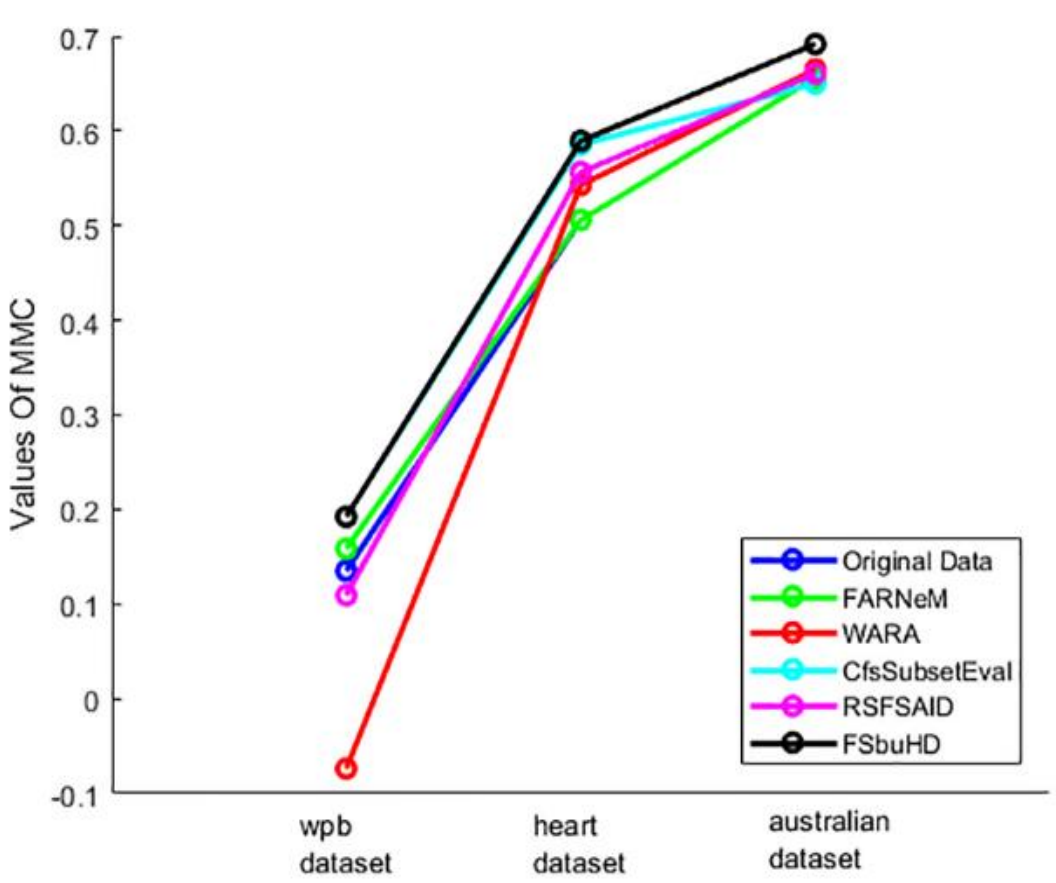

**Figure 14.** The values of MMC when using Complex Tree classifier

Overall, it can be observed that the performance and efficiency of the presented model are superior to those of previous models. This conclusion is based on the results obtained from the tables above and the analysis of the relevant results (underlined numbers) across different datasets.

## 6. CONCLUSIONS

The present study introduces a novel model for feature selection based on fuzzy rough set theory, focusing on its application within a hybrid information system. This model addresses the limitations of traditional feature selection methods, particularly in high-dimensional spaces and hybrid information systems, where conventional techniques may fall short.

The newly proposed model is named **FSbuHD**. In this model, a hybrid distance measure is employed to establish a similarity relation between objects. This similarity relation is utilized in both normal and optimistic modes, enhancing the effectiveness of feature selection.

To solve the FSbuHD model, we apply a meta-heuristic algorithm known as the black hole algorithm, which is inspired by natural phenomena. After implementing the algorithm, we present the feature selection results obtained from eight different datasets sourced from the UCI Machine Learning Repository.

The acceptability of the FSbuHD model is demonstrated through a comparison of its selected features with those derived from previously established algorithms. To evaluate the performance of the proposed model against other algorithms, three datasets were randomly selected. The performance metrics—accuracy, precision, recall, and Matthews correlation coefficient—were computed using three classification algorithms: SVM, KNN, and Complex Tree. The results indicate that the FSbuHD model outperforms previous algorithms, highlighting its effectiveness in feature selection.

For future research, several avenues can be explored to further enhance the FSbuHD model. One potential direction is to solve the model using various other meta-heuristic algorithms, allowing for a comprehensive comparison of their performance in terms of both the quantity of selected features and the quality of the results obtained. Additionally, investigating alternative equivalence relations could represent a significant advancement in the field of feature selection and warrant further study.

In conclusion, this research contributes to the growing body of knowledge in feature selection methods by introducing a robust model that effectively leverages fuzzy rough set theory within hybrid information systems. The promising results obtained from the FSbuHD model suggest its potential for wider application and further development in future studies.

## 7. ACKNOWLEDGEMENT

Data for this study come from the UCI repository.

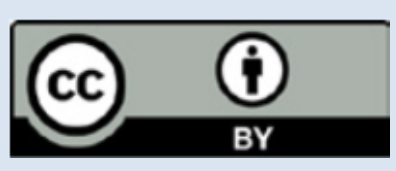

## Persian Abstract

**چکیده**

با توجه به حجم بالا، تنوع زیاد و سرعت بالای تولید داده ها، بررسی روش های انتخاب ویژگی از کلان داده کاربردها و مزایای مختلفی دارد. انتخاب ویژگی با حذف ویژگی های نامرتبط و تکراری منجر به کاهش ابعاد داده می شود که به تصمیم گیری بهینه در یک سیستم تصمیم گیری کمک می کند. یکی از ابزارهای مهم برای انتخاب ویژگی در سیستم های اطلاعات ترکیبی، نظریه مجموعه های راف فازی است. با این حال، از آنجا که این نظریه از روابط هم ارزی فازی برای انتخاب ویژگی استفاده می کند، دو مشکل مهم دارد. نخست اینکه، به دست آوردن روابط هم ارزی فازی از طریق عملیات های ممتد اشتراک در ابعاد بالا نیازمند صرف زمان و حافظه زیاد می باشد و ضعف دیگراین روش نیز تولید بسیار زیاد داده های نویزی در ابعاد بالاست. هدف و نوآوری این مقاله غلبه بر این مشکلات است. این مقاله یک مدل جدید برای انتخاب ویژگی پیشنهاد می کند. در این مدل فاصله هیبریدی بین اشیا محاسبه و سپس از آن برای به دست آوردن رابطه هم ارزی فازی استفاده می شود. در واقع، به جای حل مسئله انتخاب ویژگی بصورت مستقیم، این روش آن را به یک مسئله بهینه سازی تبدیل می کند که می توان آن را با استفاده از یکی از الگوریتمهای فرا ابتکاری مناسب حل کرد. نویسندگان، این رویکرد جدید را FSbuHD نامگذاری کرده اند. مدل جدید، بر اساس انتخاب یکی از دو رابطه هم ارزی فازی معرفی شده، دارای دو حالت عادی و خوشبینانه خواهد بود. مدل FSbuHD سپس بر روی داده های استاندارد UCI آزمایش و با سایر الگوریتمها مقایسه شده است. نتایج این تحقیق نشان می دهد که این روش یکی از کارآمدترین و مؤثرترین روشها برای انتخاب ویژگی در مقایسه با روشها و الگوریتمهای قبلی است.